\documentclass[conference]{IEEEtran}

\IEEEoverridecommandlockouts
\usepackage[pdftex]{graphicx}
\usepackage{epstopdf}
\usepackage{cite}
\usepackage{amsmath,amssymb,amsfonts}
\usepackage{algorithmic}
\usepackage{graphicx}
\usepackage{textcomp}
\usepackage{balance}
\usepackage{xcolor}
\usepackage{hyperref}
\def\BibTeX{{\rm B\kern-.05em{\sc i\kern-.025em b}\kern-.08em
    T\kern-.1667em\lower.7ex\hbox{E}\kern-.125emX}}

\begin{document}

\title{ChunkFormer: Learning Long Time Series with Multi-stage Chunked Transformer}

\author{Yue Ju$^{\ast}$, Alka Isac$^{\ast}$ and Yimin Nie $^{\ast\dag}$ \\
$^{\ast}$Global AI Accelerator, Ericsson, Montreal, Canada, \\
 }
\maketitle

\begin{abstract}
The analysis of long sequence data remains challenging in many real-world applications. We propose a novel architecture, ChunkFormer, that improves the existing Transformer framework to handle the challenges while dealing with long time series. Original Transformer-based models adopt an attention mechanism to discover global information along a sequence to leverage the contextual data. Long sequential data traps local information such as seasonality and fluctuations in short data sequences. In addition, the original Transformer consumes more resources by carrying the entire attention matrix during the training course. To overcome these challenges, ChunkFormer splits the long sequences into smaller sequence chunks for the attention calculation, progressively applying different chunk sizes in each stage. In this way, the proposed model gradually learns both local and global information without changing the total length of the input sequences. We have extensively tested the effectiveness of this new architecture on different business domains and have proved the advantage of such a model over the existing Transformer-based models.
\end{abstract}


\section{Introduction}
\noindent
Time series analysis has gained a lot of momentum in real-world use cases like telecommunication, advertisement click analysis, online education, and sales \& retail to study the underlying causes of trends or systemic patterns captured over time.  Deep learning techniques such as Recurrent Neural Network (RNN) based models are widely accepted as state-of-the-art approaches while dealing with sequential data. RNN based models such as LSTM and GRU are applied to extract the dependencies from the historical sequences to predict the likelihood of future events. However, regular RNNs are hard to train due to vanishing and exploding gradient problems making the model unstable and unable to learn global relationships between input items from long sequential data. 
\\

\noindent Transformer architecture was widely accepted as a novel approach to outperform the regular RNN based models by retrieving global information with an attention mechanism. However, for the implementation of these models, we need to specify a cut-off length before modeling, thus making the decision of the cut-off length random and subjective. In some real-world applications, the crucial information in a long sequence data is concealed locally and globally. Some important local information might reside far from the maximal cut-off sequence. For example, the utilization of mobile traffic exhibits fluctuations in short time stamps within a long time sequence. Additionally, canonical Transformers with self-attention mechanisms are computationally expensive due to the quadratic complexity of sequence length.  

\noindent
\\
To address the above issues, many models based on Transformer architecture have been proposed recently \cite{zhou2021informer,kitaev2020reformer,li2019enhancing}. The Informer model uses a Multi-head ProbSparse self-attention mechanism which achieves $\mathcal{O}(L\log L)$ in time complexity and memory usage \cite{zhou2021informer}. Reformer achieves the same time complexity by replacing dot-product attention with locality-sensitive hashing \cite{kitaev2020reformer}. LogSparseTransformer tackled forecasting problem with long time series to enhance the locality and reduce the memory bottleneck of Transformer\cite{li2019enhancing}. It introduces convolutions and LogSparse self-attention so that local context can be better incorporated into the attention mechanism. AutoFormer studies how to decompose temporal patterns from a long-time sequence by incorporating Auto-Correlation and attention mechanisms, yielding a state-of-art accuracy on multiple domains \cite{wu2021autoformer}.

\noindent
\\
The recent development on Transformer-based models in time series analysis has shown that the combination of Transformer and convolutional-style operation seem to improve the capability of a model to discover both local and global information in a long sequence. However, we found that all the existing models might not fully sustain the information for a long sequence because the application of convolutional-style computation and LogSparse attention might lose certain information for the entire sequence. Therefore, in this paper, we propose a novel structure: Multi-Stage ChunkFormer Architecture. ChunkFormer still follows the basic attention mechanism with only an encoder structure but applies Transformer attention on smaller chunks in a long sequence. The local information will be computed within each small chunk Transformer block. ChunkFormer also uses Multi-Stage blocks to chunk the local group progressively. Therefore, the proposed structure gradually learns information from local to global in different phases and retains the original length of input during the training course. Multi-Stage ChunkFormer achieves state-of-the-art accuracy on five public datasets from telecommunication, advertisement click, sales in retails, and online education.   
\\

\noindent The contributions are summarized as follows:

\begin{description}
\item[$\bullet$] ChunkFormer extracts local and global information from a long sequence data and retains the sequential length during the entire training course
\item[$\bullet$] The novel architecture overcomes the memory bottleneck challenges of Transformer architecture and outperforms the existing models on accuracy 

\item[$\bullet$] ChunkFormer achieves improvement for extremely long time series with seasonality and local fluctuations on three real-world use cases: telecommunication, advertisement click, and online education
\end{description}

\section{Related Work}

\noindent Traditional methods for time series analysis typically involve the classical approaches, and some non-stationary analysis based on ARIMA. The recent advancement of Deep Recurrent Neural Network (RNN) provides new tools for time series forecasting\cite{schmidt2019recurrent}.  These models extract the temporal dependencies from time sequences. LSTM\cite{hochreiter1997long} and GRU\cite{cho2014learning} are widely used RNNs that reach the state-of-the-art performance in many time series tasks by introducing forgetting gates that regulate the flow of information, thus overcoming short time memory issues of regular RNNs.
\\

\noindent LSTM and GRU, however, still encounter difficulties extracting global relationships hidden within the long data sequences. Attention-based RNNs\cite{bahdanau2014neural} therefore were proposed to explore the long-range dependencies between time sequences. The attention mechanism computes the attention score to identify the correlation for each item from the entire sequence of data. This advanced methodology makes it possible to capture the relationship between input items at timestamps far away that might have a tight correlation. This relationship is difficult to be extracted by regular RNNs. 
\\

\noindent Soon after the introduction of the attention mechanism, Transformers based on self-attention mechanism were proposed \cite{devlin2018bert} and have proved to show great power in sequential analysis not only in Natural Language Processing and Audio Processing, but also in Computer Vision. However, many studies show that applying self-attention to long time sequences is still computationally challenging. It takes quadratic complexity of sequential length $L$ in both memory and time. The high complexity is because the self-attention score is computed by taking all the items in the sequence as a squared matrix for each pair of items. To reduce the complexity and utilize the performance of the self-attention mechanism, researchers have proposed some improved Transformer architectures. LogSparseFormer introduces the local convolution to regular Transformer and a LogSparse attention mechanism that selects time steps following the exponentially increasing intervals \cite{child2019generating}. Reformer shows the local-sensitive hashing (LSH) attention and reduces the complexity significantly \cite{kitaev2020reformer}. Informer Models extends Transformer with KL-divergence based ProbSparse attention and also reaches $\mathcal{O}(L\log L)$\cite{zhou2021informer}. AutoFormer incorporates auto-correlation and time series decomposition techniques with Transformers and achieves state-of-the-art performances on multiple real-world use cases \cite{wu2021autoformer}.
\\

\noindent The above advancements reduce model complexity by introducing sparse attention layers. Due to that, these approaches might result in the loss of certain hidden information on long-time sequences. In this paper, we present a chunk-style Multi-Stage Transformer to reduce the complexity and improve model performance. The model takes the complete information from a long sequence and applies Multi-Stage Transformers progressively. We show that this novel model architecture provides state-of-the-art performance on multiple real-world datasets in telecommunication, online advertisement, and online education.
\section{ChunkFormer}
\noindent As mentioned in the earlier sections, handling long sequence data can be challenging due to the trade-off of capturing global information along the sequence without compromising local information contained within small segments of the sequence data. Tackling computational efficiently while managing both these aspects is another bottleneck. To address these challenges in real time data, we propose a convolution style block transformer series, ChunkFormer, that can maneuver both these challenges efficiently.

\subsection{Model Architecture}
\noindent We revamp the Traditional Transformer model to a series of small transformer blocks with repeating stages to capture the short-term and long-term dependencies among input vectors at different time stamps. The chunk style architecture captures the local patterns hidden within neighboring regions while the Multi-Stage architecture gradually captures the global information within distant vectors. As the attention mechanism is calculated for small chunks of sequence, the overall model is computationally efficient for data with large number of features and sequence lengths.
\\

\subsubsection{Chunking Traditional Transformer - Local Attention}

\hfill\break

\noindent The below figure shows the structure of traditional transformer. 

\begin{figure}[ht]
\centering
\includegraphics[width=1\linewidth]{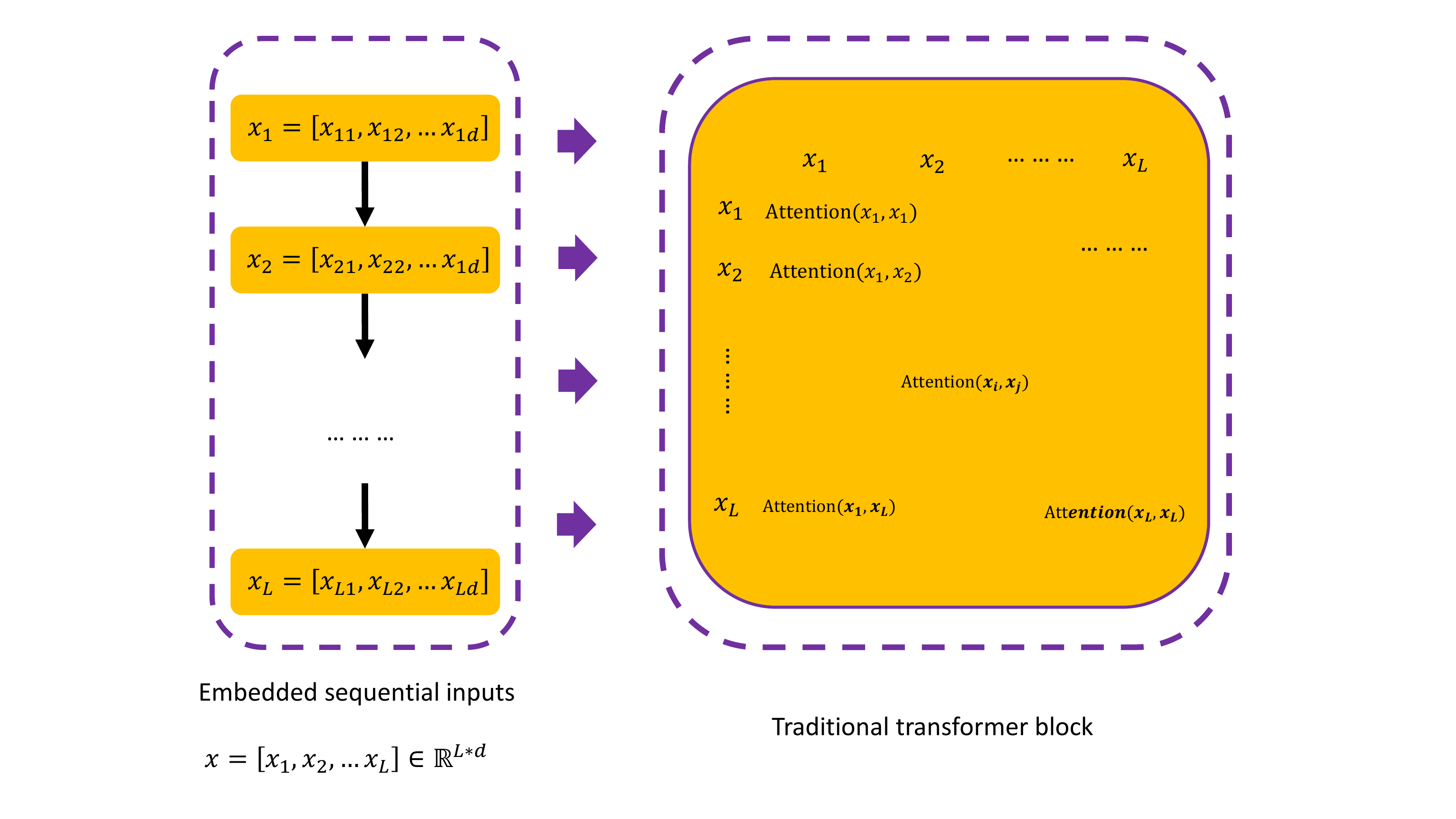}
\caption{Traditional Transformer Architecture. 
The sequential data $x=[x_1,x_2,...,x_L]$ include $L$ embedding vectors which are from time series features. Each embedding vector $x_i$ has $d$ as embedded dimension. The regular transformer calculates the attention score for each pair of embedded vectors $x_i$ and $x_j$.}
\label{fig:Fig2bis}
\end{figure}

\noindent A traditional transformer successfully captures the global relationships between distant vectors by computing attention score for all vector pairs. However, it often falls short on capturing patterns preserved within neighboring regions that are sometimes crucial to many applications in time series data. For such data, small segments in a sequence correlate more than when considering the entire input data. Moreover, for long sequence data, the attention matrix will consume more resources to maintain a large matrix, especially during the training of deep layers of the network. ChunkFormer model splits long sequence data to smaller chunks to capture these local relationships and patterns. As the attention layers are applied only within smaller sections of sequences, this approach ensures computational efficiently while working on long sequence lengths with many features.

\begin{figure}[ht]
\centering
\includegraphics[width=1\linewidth]{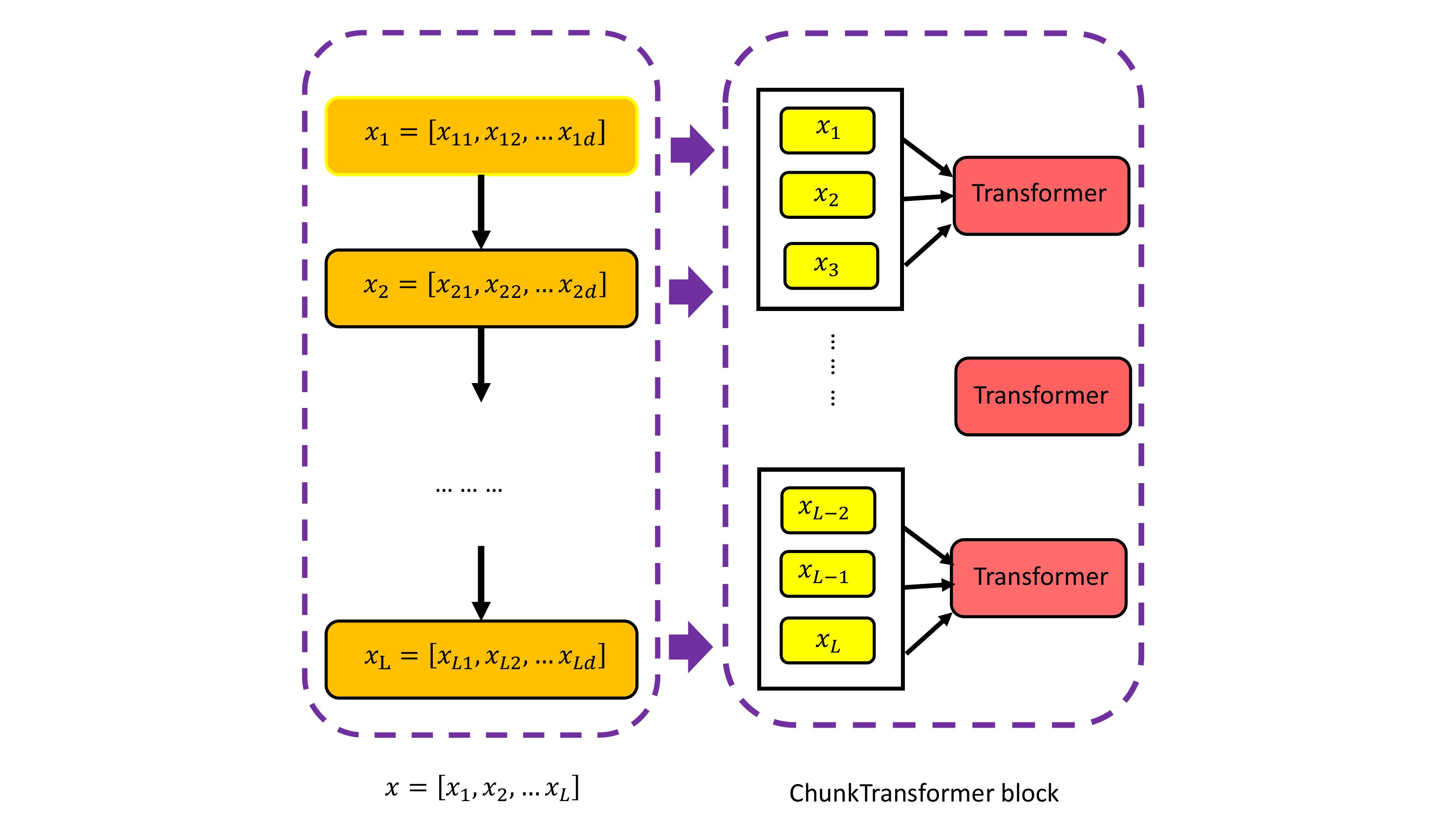}
\caption{Single-stage ChunkFormer Architecture. The entire sequential data $x=[x_1,x_2,...,x_L]$ is chunked by a chunk size $k$ (e.g., $k=3$). The attention scores are only calculated based on each smaller chunk.}
\label{fig:Fig2bis}
\end{figure}

\subsubsection{Multi-Stage Blocks - Global Attention}

The Multi-Stage block network introduced in our architecture progressively learns information between distant vector sequence, gradually learning global information hidden within the sequence. This novel architecture gradually learns from local to global information in different phases, all while retaining the length of the input sequence.

\begin{figure}[ht]
\centering
\includegraphics[width=1\linewidth]{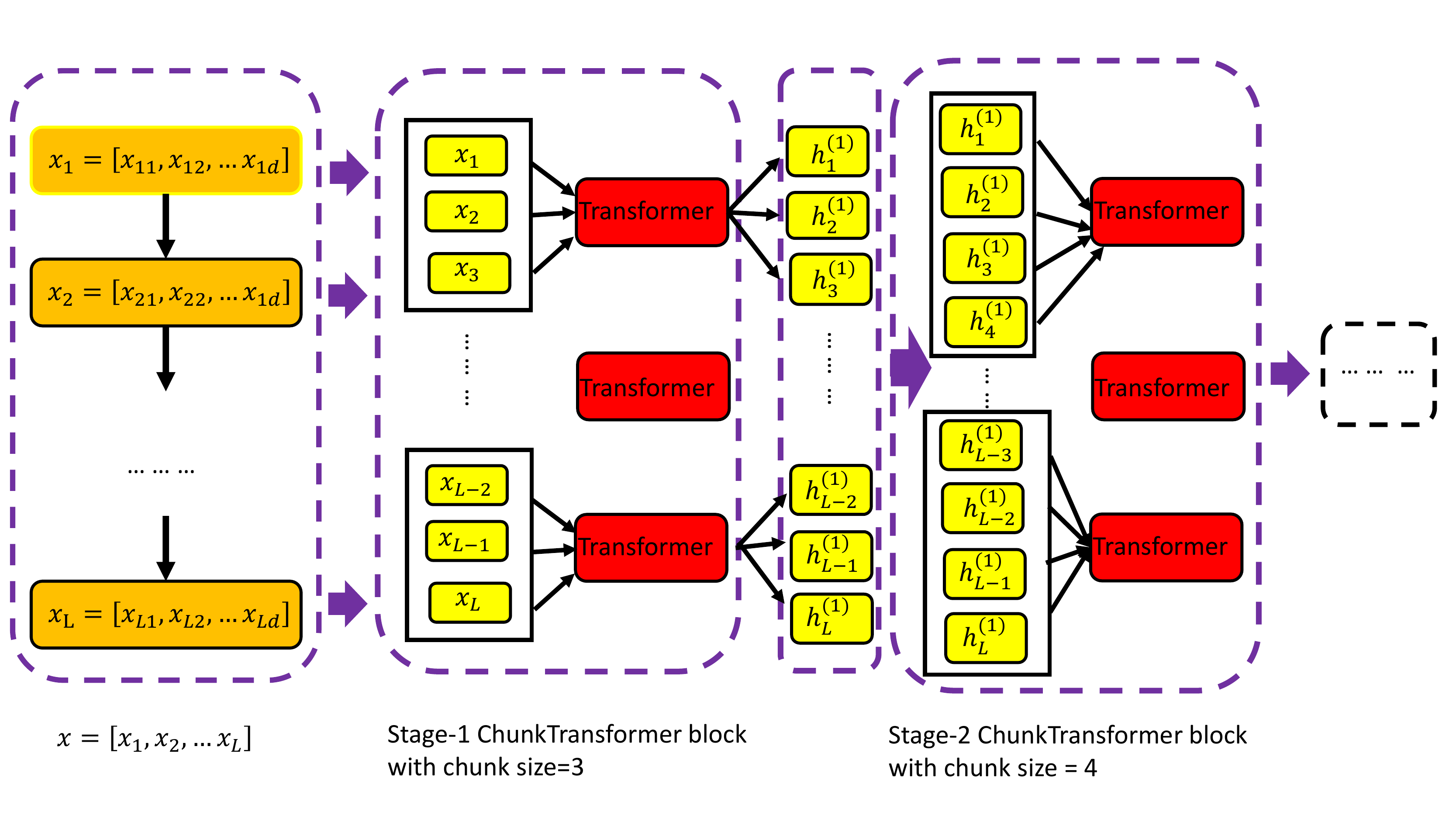}
\caption{Multi-stage ChunkFormer. A multi-stage chunkFormer will perform the calculation for attentions scores with different chunk size within each stage. For example, for a 2-stage chunkFormer with kernel size 3 and 4 for the first and the second stage.}
\label{fig:Fig2bis}
\end{figure}

\vspace{0.5em}
\begin{figure*}
\normalsize
\center
\includegraphics[width=0.9\textwidth]{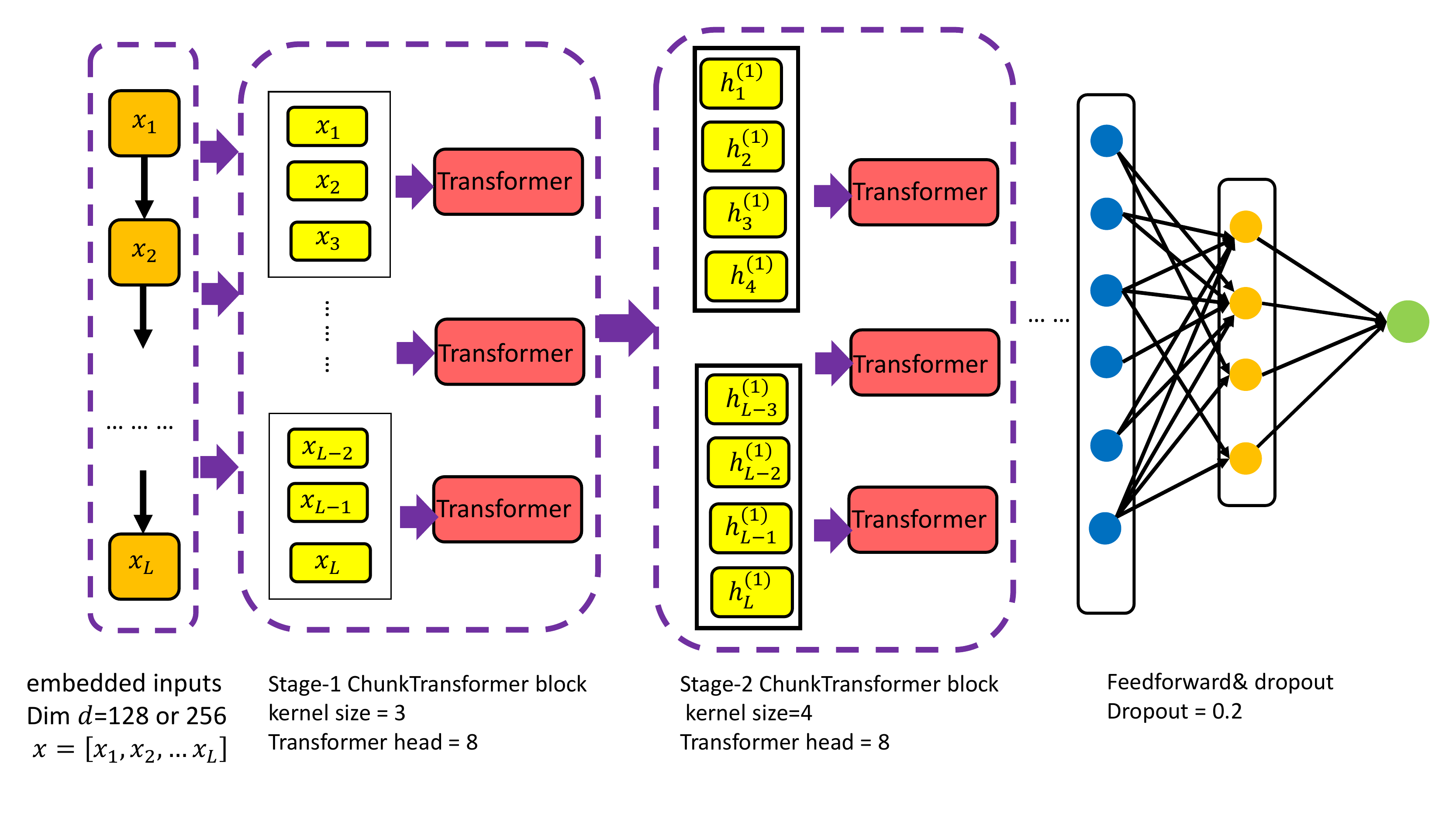}
\caption{End to End Multi-stage ChunkFormer }
\label{fig:model}
\end{figure*}

\subsection{Model Input}
\noindent The input to the encoder block of ChunkFormer is the long sequence time series data, with time factor preserved by ordering the data in the order of occurrence.\\

\noindent Consider time sequence of \textit{L} time steps \textit{X} $\in$ $\mathbb{R}$$^{L \times d}$

\noindent Each time step has dimension \textit{d} decided by the number of features associated with the time step.

\begin{equation}
\label{eq1}
\textit{X} = [\ x_1,x_2,x_3, ..., x_L ]\   
\end{equation}

\hfill

\noindent In the Stage-1 ChunkFormer block, the raw input data with \textit{L} time steps get split into shorter sequences based on chunk size. 
For chunk size \textit{k}, the input sequence will be partitioned into $B=L/K$ equal sub-regions of size \textit{k}

\begin{equation}
\label{eq2}
\begin{split}
\textit{X} & = [\ [\ x_1,x_2..x_k]\ ,[\ x_{k+1},x_{k+2},...,x_{2k}]\ ... \\ 
& [\ x_{(B-1)k+1},x_{(B-1)k+2},...,{x_{Bk}}]\ ]\ 
\end{split}
\end{equation}

\hfill

\noindent The hidden state $h^{(1)}$ for the first stage will be calculated for each of the ChunkFormer block. The below equations show the attention calculation for each block $\textit{m}\in [1,B]$.

\begin{equation} \label{eq3}
\begin{split}
T^{(1)}_m & = \textrm{Attention}([x_{(m-1)k+1},x_{(m-1)k+2}..x_{mk}]\ ) \\
& = [\ h^{(1)}_{(m-1)k+1},h^{(1)}_{(m-1)k+2},...,h^{(1)}_{mk}]\
\end{split}
\end{equation}

\hfill

\noindent The sequence length is preserved as we move across the different stages. The output of each stage is then concatenated by the outputs from small chunks and will have the same number of sequence as the input to each layer.

\hfill
\begin{equation}\label{eq4}
\begin{split}
Output_1 & = \textrm{Concatenate}([T^{(1)}_1, T^{(1)}_2,...T^{(1)}_B]) \\
& = [h^{(1)}_1,h^{(1)}_2,h^{(1)}_3,...,h^{(1)}_{L-2},h^{(1)}_{L-1},h^{(1)}_L]\
\end{split}
\end{equation}

\hfill

\noindent The next stage of ChunkFormer block will take the input of concatenated extracted features and partition it into a new different set of blocks with larger chunk size \textit{j} (\ \textit{j} $>$ \textit{k})\, then applying attention mechanism to sub-regions of the new input.

\hfill

\noindent In general, the below shows the attention calculation for kernel size \textit{j}, for the \textit{m}\textsuperscript{th} block of the \textit{s}\textsuperscript{th} stage of N-stage ChunkFormer.

\begin{equation} \label{eq5}
\begin{split}
T^{(s)}_m & = \textrm{Attention}([\ h_{(m-1)j+1}^{(s-1)},h_{(m-1)j+2}^{(s-1)},...,h_{mj}^{(s-1)}]\ ) \\
& = [\ h^{(s)}_{(m-1)j+1},h^{(s)}_{(m-1)j+2},...,h^{(s)}_{mj}]\
\end{split}
\end{equation}

\hfill

\noindent The final output after N-stage chunked transformer is given by
\hfill
\begin{equation}
\label{eq6}
\begin{split}
Output_N & = \textrm{Concatenate}([T^{(N)}_1, T^{(N)}_2,...T^{(N)}_B]) \\
& = [h^{(N)}_1, h^{(N)}_2,...,h^{(N)}_L] \
\end{split}
\end{equation}

\hfill

\noindent In such a way, we can extract more local information together with global ones in recursive stages forward. 

\hfill

\noindent The output of the final stage of \textit{n}-stage transformer will be sent through a feed-forward network to predict the target variable.

\begin{equation} \label{eq7}
Output_{final} = \textrm{Feedforward}( [\ h^{(N)}_1, h^{(N)}_2,...,h^{(N)}_L]\ )
\end{equation}

\hfill \break

\section{Experiments}
\noindent In this section, we present the performance of the novel architecture ChunkFormer on three different time series applications.\\

\noindent To evaluate the performance of our proposed ChunkFormer, we compared the model results on the selected datasets with three state-of-the-art models used in time-series applications - regular Transformer (rTransformer), LSTM, and LogSparseFormer. AUC, Macro F$_{1}$ score, and space complexity were the evaluation metrics selected for the comparison. Overall, ChunkFormer outperforms other models on these evaluation indicators. 

\subsection{Datasets}

\noindent Here is the description of the three selected datasets: 
\\ 

\begin{enumerate}

\item Content Delivery Network (CDN)\footnote{Ericsson Internal Data} dataset consists of back-end logs collected from the Ericsson network that allocates local surrogate servers according to each visit session from different IP addresses. It provides a few time series based key performance indicator (KPI) features to track the connection of serial sessions. Failure events are rare, but they can have a great impact on the quality of user experience. We aim to detect connection failures, which is essential for the system to better allocate resources and provide network services. Because of the huge amount of data in this dataset, we sampled two days of log tracking, which is approximately 5.4 million records.  
\item TalkingData (TD)\footnote{\url{https://www.kaggle.com/c/talkingdata-adtracking-fraud-detection}} contains data collected from approximately 200 million clicks over 4 days on the TalkingData platform which is one of China’s largest independent big data service platforms. Our goal here is to flag IP addresses that produce lots of clicks but never end up installing apps. In order to adapt to our resources, we extracted the first 10 million data samples from this huge dataset.
\item Online Education (OE)\footnote{\url{https://www.kaggle.com/c/riiid-test-answer-prediction}} is published by Riiid Labs, an artificial intelligence solution provider, and contains more than 100 million student interaction records. It provides all kinds of information that a complete educational application should have: the historical performance of students, the performance of other students on the same issue, metadata about the issue itself, and so on. The goal here is to accurately predict whether the student will be able to answer the next question correctly. Similar to the sampling method of the previous dataset, we selected the first 10 million rows of the OE data. 
\end{enumerate}
\noindent The classes in these samples are extremely skewed thus following the pattern that exists in most real-world time series applications.

\subsection{Implementation details} 

\noindent\textbf{\textit{Data preprocessing}}: The data preparation procedure follows the existing approaches with few modifications that make the pipeline unique and generalizable for any kind of data. 
\\

\noindent The Below highlights the unique methodology we adopted to transform the input data to a model ingestible format.
\\

\begin{enumerate}
\item We performed a standard data cleaning and customized feature engineering step to prepare the input features of the data. The feature engineering engine of our proposed framework can generalize the different types of input. It encodes classification and continuous input features to fit the ChunkFormer framework. Compared with the traditional Transformer framework, this unique engine has a significant difference. Categorical features are encoded using conventional embedding vectors, and numerical features are discretized by precise normalization and rounded to quasi-integers. \\

For example, if the maximum and minimum values of a numeric vector x are 3.752 and 0.012, respectively, the normalized vector will become a series integer [3752,...,12] with a precision of 0.001. These integers are then converted to categorical features and presented as embedding. If the discretization of numerical values results in a large embedding vector, the numbers undergo bucketing, automatically converting the numerical features to embedding like categorical features. 
\\

\item After data cleaning and feature engineering, we performed a data grouping operation to generate sequential data.\\  

For the CDN dataset, we grouped the data according to the IP of the network distribution server. For TD and OE data, the IP address and user ID columns were used respectively to group the data. To avoid data leakage, we arranged the time-sequenced input features in chronological order within each group. We further cleaned the data by removing any groups with the number of records lower than the set threshold of 2. 
\\

\item To train and evaluate the models, we split the datasets into training, validation, and test data sets according to groups. In the CDN data, the training, validation, and test sets contained 80,000, 2,000, 2,019 groups respectively. For TD data, we generated 50,000 groups for training, 10,000 for validation, and 8,740 for testing. The OE data had 20,000 user ID groups in the training set, 10,000 user ID groups in the validation set, and 3,000 user ID groups in the testing set.
\end{enumerate}

\vspace{0.5em}
\begin{figure*}
\normalsize
\center
\includegraphics[width=0.9\textwidth]{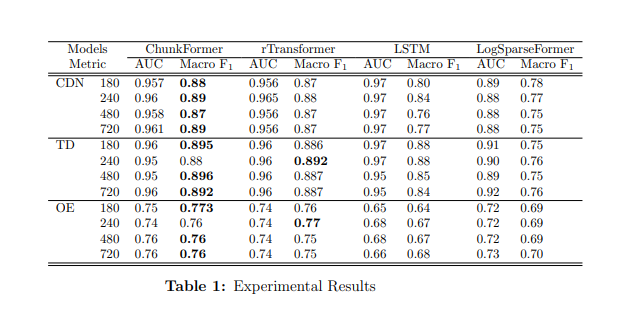}
\label{fig:model}
\end{figure*}

\noindent\textbf{\textit{Model settings}}: The Transformer, LSTM, and LogSparseFormer models were selected as baseline models to compare and evaluate the performance of our proposed model on the below performance metrics: AUC, Macro F$_{1}$ score. \\

\noindent The same hyper-parameters were set for all the deep models to have a consistent baseline for comparison: ADAM optimization algorithm\cite{adam} was used for the gradient based optimization of stochastic objective function. Sigmoid was used as the activation function. The loss function was set to BCEWithLogitsLoss. The learning rate of CDN and OE datasets was initialized to 5e-4, and the learning rate of TD data was initialized to 1e-5 for all models. The epoch number of CDN was set to 20, TD was set to 30, and OE was set to 10. All models were implemented in PyTorch and executed on a single Tesla P100 16 GB GPU.  For all model experiments of 2-stage ChunkFormer, the sequence length was set to [180, 240, 480, 720]. 

\subsection{Experimental Results}

\noindent The experimental results on the baseline models and the proposed models are shown in Table 1.  
\newline
\newline
Below are the main conclusions from the experiments: 

\begin{enumerate}
\item In general, ChunkFormer performs better than the other baseline models on Macro F$_{1}$ score, which is a more superior evaluation indicator than AUC for unbalanced data. Although the AUC value of most of the models is similar, the value of Macro F$_{1}$ score is quite different. All three experimental datasets have a high degree of imbalance between the positive and negative samples. This imbalance limits the model's ability from improving the Macro F$_{1}$ score. After the Macro F$_{1}$ score reaches a certain level, even a 1\% improvement is a game-changer in real-world applications. ChunkFormer does well in pushing the limits of F$_{1}$ scores compared to other models. For the CDN dataset, all the experimental results of ChunkFormer's Macro F$_{1}$ score are superior to those of the other baseline models for all sequence lengths. For the TD dataset, the performance of ChunkFormer is superior to that of the regular Transformer model except for sequence length 240. Similar is the case for OE experimental performance. 
\newline
\item ChunkFormer’s performance between different sequence lengths is more stable than the traditional machine learning method LSTM. For the CDN dataset, the difference between ChunkFormer's maximum and minimum Macro F$_{1}$ scores is approximately 0.02. In particular, LSTM has a difference of 0.08 between the extreme values of Macro F$_{1}$ scores. For the TD dataset, ChunkFormer has a difference of 0.02, while LSTM has a difference of 0.04 between the maximum and minimum values of Macro F$_{1}$. For the OE dataset, the difference between the maximum and minimum values of Macro F$_{1}$ score on LSTM is three times that of ChunkFormer.  
\newline
 
\item
We have observed that 2-3 stage chunkFormer is able to reach the state-of-the-art performance for all experimental datasets we have used in this paper. Since chunkFormer only takes smaller chunks for the calculation of Transformer, it reduces the space complexity from $\mathcal{O}(L^2)$ to $\mathcal{O}(kL)$ with $k<<L$. For example, assume we take N-stage Chunkformer with kernel size $k_1,k_2,...,k_N$, we apply chunked transformer recursively by overwritting the previous stage's hidden states $h^{(s)}_1,h^{(s)}_2,...,h^{(s)}_L$, therefore, the overall space complexity is $\mathcal{O}(k_NL)$. Since the kernel size on the last stage $K_N<<L$, the proposed model uses less computational resources than regular transformer. For the most recent advanced models such as LogSparseFormer with complexity $\mathcal{O}(L\log L)$, the proposed ChunkFormer is superior for longer sequences. For example, for a long sequence with $L\geq256$, assume the kernel size $K_N\leq 8$ on the last stage, the space complexity of ChunkFormer $\leq 8L$ which is smaller than the space complexity of LogSparseFormer.

\end{enumerate}

\subsection{Model Analysis}
\noindent In this section, we will analyze the architectural novelty that worked well for our proposed model.
\\

\begin{description}

\item[$\bullet$] ChunkFormer inherits the advantages of the Transformer model class and makes improvements to this framework. It combines Transformer and CNN-style structures on long sequence vectors. CNN-style operations do well in associating the information with local areas, while Transformers have the strength in focusing on global attention. In contrast, the traditional Transformer only calculates the global attention score of the entire input sequence and fails in capturing the local interactive information in the long sequence data. In addition, the LSTM-like model does not perform well when trained on long data sequences that need to learn long-term time dependence. \\

\item[$\bullet$] ChunkFormer benefits from dividing the input sequence of each stage into different sub-regions (convolution-like block). The attention matrix is calculated for each small sub-regions of input instead of the entire input sequence block. This mechanism improves the time and space efficiency of the model without compromising accuracy.  \\

The trade-off between accuracy and operational efficiency adopted by ChunkFormer can be a possible reason for the under-performance of some indicators for some sequence lengths. ChunkFormer consumes fewer resources than all baseline models making it most effective and practical in real-world applications. The random selection of block chunk size can also be another contributing factor for the reduced accuracy. 
\end{description}

\section{Conclusion}
\noindent In this paper, we proposed, implemented, and evaluated a novel Transformer based architecture, ChunkFormer, that uses a combination of Transformer attention mechanisms in a convolution-like structure on long sequence vectors. We have proved that the proposed model can capture both the local and global information hidden within the timestamps rendering it extremely valuable in real-world applications.  Our model meets the memory bottlenecks of the general Transformer based models, thus making it ideal in practical time series applications where real-time prediction is crucial for any decision making. ChunkFormer significantly improves the Macro F$_{1}$ value of detecting rare events in many business scenarios, from telecommunication and fraudulence to online education. In future studies, we will focus on improving the model performance by tuning the hyper-parameters like chunk size, learning rate, etc.


\balance

\bibliographystyle{IEEEtran}
\bibliography{references}

\begin{thebibliography}{10}
\expandafter\ifx\csname url\endcsname\relax
  \def\url#1{\texttt{#1}}\fi
\expandafter\ifx\csname urlprefix\endcsname\relax\def\urlprefix{URL }\fi
\expandafter\ifx\csname href\endcsname\relax
  \def\href#1#2{#2} \def\path#1{#1}\fi

\bibitem{zhou2021informer}
H.~Zhou, S.~Zhang, J.~Peng, S.~Zhang, J.~Li, H.~Xiong, W.~Zhang, Informer:
  Beyond efficient transformer for long sequence time-series forecasting, in:
  Proceedings of AAAI, 2021.

\bibitem{kitaev2020reformer}
N.~Kitaev, {\L}.~Kaiser, A.~Levskaya, Reformer: The efficient transformer,
  arXiv preprint arXiv:2001.04451 (2020).

\bibitem{li2019enhancing}
S.~Li, X.~Jin, Y.~Xuan, X.~Zhou, W.~Chen, Y.-X. Wang, X.~Yan, Enhancing the
  locality and breaking the memory bottleneck of transformer on time series
  forecasting, Advances in Neural Information Processing Systems 32 (2019)
  5243--5253.

\bibitem{wu2021autoformer}
H.~Wu, J.~Xu, J.~Wang, M.~Long, Autoformer: Decomposition transformers with
  auto-correlation for long-term series forecasting, arXiv preprint
  arXiv:2106.13008 (2021).

\bibitem{schmidt2019recurrent}
R.~M. Schmidt, Recurrent neural networks (rnns): A gentle introduction and
  overview, arXiv preprint arXiv:1912.05911 (2019).

\bibitem{hochreiter1997long}
S.~Hochreiter, J.~Schmidhuber, Long short-term memory, Neural computation 9~(8)
  (1997) 1735--1780.

\bibitem{cho2014learning}
K.~Cho, B.~Van~Merri{\"e}nboer, C.~Gulcehre, D.~Bahdanau, F.~Bougares,
  H.~Schwenk, Y.~Bengio, Learning phrase representations using rnn
  encoder-decoder for statistical machine translation, arXiv preprint
  arXiv:1406.1078 (2014).

\bibitem{bahdanau2014neural}
D.~Bahdanau, K.~Cho, Y.~Bengio, Neural machine translation by jointly learning
  to align and translate, arXiv preprint arXiv:1409.0473 (2014).

\bibitem{devlin2018bert}
J.~Devlin, M.-W. Chang, K.~Lee, K.~Toutanova, Bert: Pre-training of deep
  bidirectional transformers for language understanding, arXiv preprint
  arXiv:1810.04805 (2018).

\bibitem{child2019generating}
R.~Child, S.~Gray, A.~Radford, I.~Sutskever, Generating long sequences with
  sparse transformers, arXiv preprint arXiv:1904.10509 (2019).

\bibitem{adam}
J.~B. Diederik P.~Kingma, Adam: A method for stochastic optimization,
  arXiv:1412.6980 (2014).

\end{thebibliography}

\end{document}